\documentclass{article}

\usepackage{arxiv}

\usepackage[utf8]{inputenc} % allow utf-8 input
\usepackage[T1]{fontenc}    % use 8-bit T1 fonts
\usepackage{hyperref}       % hyperlinks
\usepackage{url}            % simple URL typesetting
\usepackage{booktabs}       % professional-quality tables
\usepackage{amsfonts}       % blackboard math symbols
\usepackage{nicefrac}       % compact symbols for 1/2, etc.
\usepackage{microtype}      % microtypography
\usepackage{lipsum}
\usepackage{graphicx}
\graphicspath{ {./images/} }

\usepackage{graphicx, booktabs, amsmath, cite, xcolor}

\DeclareMathOperator*{\sens}{SENS}

\DeclareMathOperator*{\sensavg}{AVG-Sensitivity}
\DeclareMathOperator*{\sensmax}{MAX-Sensitivity}

\DeclareMathOperator*{\mae}{MAE}
\DeclareMathOperator*{\mse}{MSE}
\DeclareMathOperator*{\ssin}{ssin}

\DeclareMathOperator*{\gini}{I_g}

\title{Meta-evaluating stability measures: MAX-Senstivity \& AVG-Sensitivity}

\author{
 Miquel Miró-Nicolau \\
  UGiVIA Research Group \\
  and Laboratory for Artificial Intelligence Applications (LAIA@UIB) \\
  University of the Balearic Islands \\ 
  Dpt. of Mathematics and Computer Science, 07122 Palma (Spain) \\
   \And
   Antoni Jaume-i-Capó \\
  UGiVIA Research Group \\
  and Laboratory for Artificial Intelligence Applications (LAIA@UIB) \\
  University of the Balearic Islands \\ 
  Dpt. of Mathematics and Computer Science, 07122 Palma (Spain) \\
  \And
 Gabriel Moyà-Alcover \\
  UGiVIA Research Group \\
  and Laboratory for Artificial Intelligence Applications (LAIA@UIB) \\
  University of the Balearic Islands \\ 
  Dpt. of Mathematics and Computer Science, 07122 Palma (Spain) \\
}

\begin{document}
\maketitle
\begin{abstract}
The use of eXplainable Artificial Intelligence (XAI) systems has introduced a set of challenges that need resolution. The XAI robustness, or stability, has been one of the goals of the community from its beginning. Multiple authors have proposed evaluating this feature using objective evaluation measures. Nonetheless, many questions remain. With this work, we propose a novel approach to meta-evaluate these metrics, i.e. analyze the correctness of the evaluators. We propose two new tests that allowed us to evaluate two different stability measures: AVG-Sensitiviy and MAX-Senstivity. We tested their reliability in the presence of perfect and robust explanations, generated with a Decision Tree; as well as completely random explanations and prediction. The metrics results showed their incapacity of identify as erroneous the random explanations, highlighting their overall unreliability.
\end{abstract}

% keywords can be removed

\section{Introduction}

Machine Learning model have become the \textit{de facto} standard solution in multiple field. This trend is greatly increased with the first functional Deep Learning models proposed by Krizhevsky \emph{et al.}~\cite{krizhevsky2012imagenet}. These models are characterized for a large internal complexity that allowed to learn high level concepts, however, this high complexity is also the reason of the so called 'black-box' problem. 

Black box models according to Guidotti~\cite{guidotti_evaluating_2021} ``provide hardly any mechanisms to explore and understand their behavior and the reasons underlying the decisions taken''. This fact is specially problematic for a sensitive field as medical praxis, where the lack of transparency (the opposed to a black box) make “clinicians uncertain about the signs of diagnosis”~\cite{chaddad_survey_2023}. With the goal to fix the black-box issue, eXplainable Artificial Intelligence (XAI) emerged, proposing to ``make a shift towards more transparent AI. It aims to create a suite of techniques that produce more explainable models whilst maintaining high performance levels''~\cite{adadi2018}. 

%According to Adadi and Berrada~\cite{adadi2018} exists four main reason behind the usage of XAI techniques: verification, upgrade, discover, and legal aspects. Recently, the European Commision approved the Artificial Intelligence (AI) act aiming to regulate the usage of this technology in the European territory. This law proposed a risk assessment for any AI application, dividing them into four different categories: Unacceptable risk, High risk, limited risk,  Minimal or no risk. The application within the first category, unacceptable risk, are forbidden. Application with minimal or no risk are not regulated. However, both High and Limited risk applications must fulfil a set o requirements: “appropriate human oversight measures to minimise risk” and “high level of robustness, security, and accuracy”. In summary, the need for transparency, and reliable XAI.

Multiple authors have reviewed the XAI field from different points of views \cite{eitel2019testing, VanderVelden2021, miro2022evaluating, chaddad_survey_2023, hohl_opening_2024}. Various conclusions can be dawned for these reviews: the existence of completly different approaches \cite{ribeiro2016should, zhou2016learning, zeiler2014visualizing, bach2015pixel, chattopadhay2018grad, simonyan2014very, muddamsetty2022visual} to identify the reason behind the predictions, known as explanation or interpretation; this large diversity on explanation methods have provoked a lack of consensus,  Krishna \emph{et al.}~\cite{krishna_disagreement_2022} identified and analysed this lack of consensus and called it \textit{the disagreement problem}. 

Adebayo \emph{et al.}~\cite{adebayo2018sanity} also identified this lack of consensus and proposed a set of sanity checks to be fulfilled by a correct explanation. These sanity checks are part of a  new research topic that aimed to measure, objectively, different qualities of the explanation. This objectivity constrasts with the \textit{ad-hoc} evaluation done by most authors, as stated by Miller~\cite{miller2019explanation} ``most of the work about explainability relies on the authors’ intuition, and an essential point is to have metrics that describe the overall explainability with the aim to compare different models regarding their level of explainability". Different authors \cite{Hoffman2018, mohseni2021multidisciplinary} reviewed the explainability features to measure. These features have been studied with different level of depth, for example fidelity had been largely studied from different points of view: objective metrics~\cite{Samek2017, Alvarez-Melis2018, yeh2019fidelity, bhatt2020evaluating}, sanity checks for these metrics~\cite{tomsett2020sanity}, synthetic benchmarks~\cite{arras2020ground, miro2023novel, mamalakis2022investigating, guidotti_evaluating_2021}, and meta evaluations~\cite{tomsett2020sanity, hedstrom2023, miro2024comprehensive}. This in depth-analysis of fidelity contrast with the shallowness in other important features as stability.

Stability, also known as robustness, is according to Alvarez-Melis \emph{et al.}~\cite{Alvarez-Melis2018} the expectation that if the data is slightly modified therefore the explanation of this modified data should be similar to the original explanation. Multiple authors have proposed stability metrics \cite{Alvarez-Melis2018, agarwal2022rethinking, yeh2019fidelity}, producing a novel disagreement on how to calculate this feature. Hedström \emph{et al.}~\cite{hedstrom_meta-evaluation_2023} proposed a method to meta evaluate XAI measures. These authors introduced a set of conditions to be fulfilled by correct measurements, and make them into a set of numerical metrics. They applied these metrics to 10 different XAI measures, two of them of robustness (MAX-Sensitivity~\cite{yeh2019fidelity} and Local Lipschitz Estimate~\cite{Alvarez-Melis2018}), without clear conclusions. The proposal of Hedström \emph{et al.}~\cite{hedstrom_meta-evaluation_2023} can be categorised as axiomatic evaluations because they establish a set of axioms and assess whether the metrics align with them.

This work aimed to improve the knowledge on stability metrics via the meta evaluation of the existing proposals. Our goal is to surpass the existing axiomatic approaches (Hedström \emph{et al.}~\cite{hedstrom_meta-evaluation_2023}) using \textit{a priori} information of the explanations. On one hand, we used a transparent model that allowed us to know the exact explanation of the prediction, particularly a Decision Trees~\cite{breiman1984classification}. This approach is similar to the meta-evaluation done by Miró-Nicolau \emph{et al.}~\cite{miro2024comprehensive} for fidelity metrics. On the other hand we used random noise both for the ``explanation'' and for the  ``prediction'', consequently provoking a lack of robustness. In the first case any correct robustness measure must be always perfect, while in the second one must show the contrary. {\color{black} In contrast to the only existing stability meta-evaluation approach, Hedström \emph{et al.}~\cite{hedstrom_meta-evaluation_2023}, our proposal is completely verifiable, without depending on any novel axiom but instead on well-defined scenarios on which the actual robustness is known. However, our approach is not suitable to be used in real scenarios, only working as a benchmark. This benchmark allowed us to discard erroneous approaches, because if are not working the simpler scenarios are not reliable also in real and complex scenario, but not to identify the the metrics that correctly worked in the simple scenario but failed in the real one.}

In this paper, we propose evaluating two robustness metrics: Average and Maximum Sensitivity, both proposed by Yeh \emph{et al.}~\cite{yeh2019fidelity}. Because we have as a prior knowledge the real robustness of the explanations, if any metric result differs from this value we will be able to detect the erroneous behaviour of the metrics.

% With this work we aimed to improve the knowledge on stability metrics via the meta evaluation of the existing proposals. Our goal is to surpass the existing axiomatic approaches (Hedström \emph{et al.}~\cite{hedstrom_meta-evaluation_2023}). To do so we used a transparent model, Decision Trees~\cite{breiman1984classification}, that allowed us to know the exact explanation of the prediction. This approach is similar to the meta-evaluation done by Miró-Nicolau \emph{et al.}~\cite{miro2024comprehensive} for fidelity metrics. We evaluate four different metrics. Because we know that the explanations are perfect any metric result different the perfection will be considered a clear sign of some mistake in the metrics. 

The rest of the paper is organized as follows: in the following section we analysed the two robustness  metrics used, in Section \ref{sec:method} we present the methodology to meta-evaluate them, in Section \ref{sec:experimental_setup} we present the experimentation setup that allowed us the meta evaluation process, in Section \ref{sec:results} we show and discuss the results obtained and in Section \ref{sec:concl} we discussed an overall conclusion of the results and the future work.

\section{Robustness Measures} \label{sec:state-of-the-art}

The objective evaluation of XAI features is a complex problematic because of the lack of a GT to compare with. This limitation is discussed by Hedström \emph{et al.}~\cite{hedstrom_meta-evaluation_2023} and refereed to it as the \textit{Challenge of Unverifiability}. For this reason XAI metrics, including robustness metrics, are always based on some assumption about the behaviour of the model and the feature to analyse itself. 

Hedström \emph{et al.}~\cite{hedstrom_meta-evaluation_2023} presented the first work that aimed to evaluate the quality of robustness metrics. These authors compared, via an axiomatic approach, the proposals from Yeh \emph{et al.}~\cite{yeh2019fidelity} and Alvarez-Melis \emph{et al.}~\cite{Alvarez-Melis2018}, without clear results. This similar results are coherent with the work from Yeh \emph{et al.}~\cite{yeh2019fidelity} that states that their approach ``is closely relate'' to Alvarez-Melis \emph{et al.}~\cite{Alvarez-Melis2018}. {\color{black} Particularly, both authors proposed to calculate the robustness using the sensitivity of a function via its gradient $w.r.t.$ of the input.} For this reason in our work we only compare the proposals from Yeh \emph{et al.}~\cite{yeh2019fidelity}.

% Alvarez-Melis \emph{et al.}~\cite{Alvarez-Melis2018} proposed to borrow from calculus the notion of function stability, in particular these authors based their metric in the Lipschitz continuity. A function $f: \mathbb{R}^n \rightarrow \mathbb{R}^m$, has Lipschitz continuity if exists a constant $L > 0$ that satisfies Equation \ref{eq:lip}:

% \begin{equation}
%     || f(x) - f(y) || < L \cdot || x - y ||, \forall x,y \in \mathbb{R}^n
%     \label{eq:lip}
% \end{equation}

% Equation \ref{eq:lip} is a global definition of continuity, however as stated by Alvarez-Melis \emph{et al.}~\cite{Alvarez-Melis2018}: "in the context of interpretability, such a notion is not meaningful since there is no reason to expect explanation uniformity for very distant inputs". For this reason these authors proposed to define an environment around an input data $x_i$ as follows:

% \begin{equation}
%     \mathcal{N}_{\epsilon}(x_i) = {x_j \in X | ||x_i - x_j || \leq \epsilon},
%     \label{eq:n_lip}
% \end{equation}

% where $\mathcal{N}_{\epsilon}(x_i)$ is a ball of radius $\epsilon$ around $x_i$, and $X = {x_i}^n_{i=1}$ denote a sample of input examples. 

% Finally the robustness metric proposed by these authors is the approximation $\tilde{L}_X$ from equations \ref{eq:lip} and \ref{eq:n_lip}, over test set $X$:

% \begin{equation}
%     \tilde{L}_X(x_i) = \argmax_{x_j \in \mathcal{N}_\epsilon(x_i) \leq \epsilon} \frac{|| f(x_i) - f(x_j) ||_2}{||x_i - x_j||_2},
%     \label{eq:lip_approx}
% \end{equation}

% where $f(x_i)$ is the explanation for input $x_i$

Yeh \emph{et al.}~\cite{yeh2019fidelity} made two proposals to calculate the explanation robustness. These authors used the sensitivity of a function via its gradient $w.r.t.$ of the input to obtain the robustness. Yeh \emph{et al.}~\cite{yeh2019fidelity} proposed to calculated a local version of this sensitivity, $\sens$, and used the maximum and average around this locality, defined by a radius $\epsilon$ and sampled with the Monte-Carlo algorithm, as robustness measures:

\begin{equation}
    \sensmax(x_i) = \max_{||x_j - x_i|| < \epsilon} || f(x_j) - f(x_i) ||,
    \label{eq:sens_max}
\end{equation}

\begin{equation}
    \sensavg(x_i) = \frac{\sum_{x_j \in S_{x_i}^\epsilon} || f(x_j) - f(x_i)||}{|S_{x_i}^\epsilon|},
    \label{eq:sens_avg}
\end{equation}

where $f(x)$ indicated the explanation for the prediction $x$; $|| ||$ the \textit{Frobenius norm}; and

\begin{equation}    
    S_{x_i}^\epsilon = \{x_j \text{ sampled using Monte-Carlo algorithm } \mid ||x_j - x_i|| < \epsilon\}
\end{equation}

In the next section we present the methodology to evaluate these measures and detect their overall performance.

\section{Method} \label{sec:method}

We proposed two tests to evaluate the performance of robustness metrics: Perfect Explanation Test (PET) and Random Output Test (ROT). Both tests are defined depending on the context and the nature of the explanations evaluated.

To be able to define this different tests we set a methodology for robustness as follows.

Let $h: \mathcal{X} \rightarrow \mathcal{Y}$ be a model. It maps instances $x \in \mathcal{X}$, the set of possible input data, to their respective outputs $y \in \mathcal{Y}$, where $\mathcal{Y}$ denotes the set of all ground truths for $\mathcal{X}$. We write $h(x)=y$ to represent the AI result for a particular input $x \in \mathcal{X}$. 

These $h$ models can be either transparent or opaque. The main difference is the availability of an explanation $e_x \in \mathcal{E}$ for an input $x$. Explanation are functions $f: \mathcal{X} \times \mathcal{Y} \rightarrow \mathcal{E}$, and XAI methods approximate them, $\hat{f}(x) \approx e_x$. We denote the results of $\hat{f}(x)$ as $\hat{e} \in \hat{\mathcal{E}}$. Finally be $r: \mathcal{E} \times \mathcal{X} \rightarrow R$ the robustness of the explanations $\mathcal{E}$ for input $\mathcal{X}$. Because of the Challenge of Unverifiability~\cite{hedstrom_meta-evaluation_2023}, we did not dispose of $\mathcal{E}$ but the approximation $\hat{\mathcal{E}}$, we can not calculate directly this $r$, therefore we approximate it with the robustness metrics analysed in the previous section, refereed as $\hat{r} \in \hat{\mathcal{R}}$, where $\hat{\mathcal{R}}$ is the set of all possible robustness metrics. This approximation can be a source of problems and is the element that must be measured.

\subsection{Perfect Explanation Test (PET)}

With this test we aimed to analyse the behaviour of robustness measures in the presence of perfect explanations. To do so we used a transparent model that allowed us to have a perfect explanation $\mathcal{E}$ for any prediction.

Taking into account that using a transparent model allowed us to dispose of the real $\mathcal{E}$ and $\mathcal{X}$, if we use a robustness metric $\hat{r}$ that is correct it must have perfect results. {\color{black} The PET test produces that the robustness metric from the original formulation, $\hat{r}: \hat{\mathcal{E}} \times \mathcal{X} \rightarrow R$, becomes the one seen in Equation \ref{eq:pet}}

\begin{equation}
    \hat{r}: \mathcal{E} \times \mathcal{X} \rightarrow R.
    \label{eq:pet}
\end{equation}

{\color{black} Therefore, the $\hat{r} = r$ must be true after applying Equation \ref{eq:pet}, and have a perfect result such as $r = 1$. Any erroneous metric $\hat{r}$ will produce a difference between the metric and real robustness value, $\hat{r} \not\approx r$, in this case $r = 1$.}

\subsection{Random Output Test (ROT)}

With this test we aimed to analyse the behaviour of robustness metrics in the presence of random explanations and predictions. 

{\color{black} Particularly, we proposed to use Gaussian Noise as explanation, and uniform noise as the model prediction. The resulting XAI system robustness is converted from the original Equation $\hat{r}: \mathcal{E} \times \mathcal{X} \rightarrow R$ to Equation \ref{eq:rot}}.

\begin{equation}
    \hat{r}: \mathcal{N}(\mu,\,\sigma^{2}) \times \mathcal{X} \rightarrow R,
    \label{eq:rot}
\end{equation}

{\color{black} Therefore, and due to the presence of the Gaussian noise} the ``model'' is not robust, $r=0$, where $0$ value indicated a completely \textit{unrobust} explanation. Consequently, any {\color{black} robustness metric} $\hat{r} \neq 0$ is clearly an erroneous results. 

In the following section we will define an experimental setup to evaluate the reviewed robustness metrics following the methodology proposed in this section, i.e. we will check whether $\hat{r} \approx r$ is true or not.

\section{Experimental setup} \label{sec:experimental_setup}

The experimental setup defined in this section was designed to evaluate the robustness metrics analysed in the Section \ref{sec:state-of-the-art} using the methodology proposed in Section \ref{sec:method}. 

\subsection{AI Model}

We evaluated MAX-Senstivity and AVG-Sensitiviy with two different tests. The first one, the Perfect Explanation Test is based on the usage of a transparent model, in this work we used a Decision Tree. 

Decision tree is a supervised, transparent AI model, internally shaped following a tree structure. Its purpose is to predict a specific outcome by learning if/then rules from provided data~\cite{breiman1984classification}. While typically used with tabular data, we can adapt it for images. We do this by flattening each image into a single vector, treating individual pixels as features.

The usual explanations from these models are global ones, with a single explanation for the whole model instead of explaining the decision for one input. The metrics analysed in this work, in contrast, were designed to analyse local explanations. To obtain a local explanation, we developed a new and simple algorithm. Because decision trees use the chosen path from root to leaf to make predictions, and each step in this path relies on analyzing a single feature, we consider all of these features to be significant in contributing to the final outcome. To determine the degree of this contribution, we used the Gini impurity criterion.

Gini impurity criterion is used to train Decision Trees. This criterion measure how pure (homogeneous) a node in the tree is with respect to the target variable (class). See Equation \ref{eq:gini} for the exact calculation.

\begin{equation}
    \gini(t) = 1 - \sum^J_{i=1} p_i^2,
    \label{eq:gini}
\end{equation}

where $t$ is the node considered, $J$ the set of all classes, and $p_i$ the proportion of data points in node $t$ that belong to class $i$. Using this criterion we obtained the importance of each node as the  difference of Gini impurity before and after the split. Because each node consider only one features this difference can be used as a proxy for the importance of the feature itself, as a more important feature provoked a larger improvement of the data split. The importance calculation can be seen at Equation \ref{eq:import}

\begin{equation}
    R_{i,j} = \left| \: \gini(t_{i-1}) - \gini(t_i) \: \right|,
    \label{eq:import}
\end{equation}

where $R_{i,j}$ is the relevance of node $t_i$, and therefore for the feature $j$ used in this node, $t_{i-1}$ is the father node of $t_i$. Finally and because multiple nodes, $\mathcal{J} : \{ \forall R_{x, y} | y = j\}$, can use the same feature $j$ the relevance of this feature is calculated as the summation of all individual importance of nodes in $\mathcal{J}$, as follows:

\begin{equation}
    R_j = \sum^{|\mathcal{J}|}_{i=1} R_{i, j},
    \label{eq:import_complete}
\end{equation}

\begin{figure}[htbp]
	\centering
     	\includegraphics[width=0.24\textwidth]{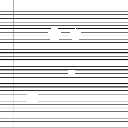}
    	\hfil
    	\includegraphics[width=0.24\textwidth]{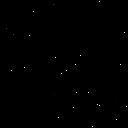}
        \hfil
    	\includegraphics[width=0.24\textwidth]{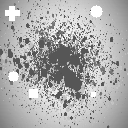}
    	\hfil
        \includegraphics[width=0.24\textwidth]{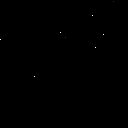} \\ 
     \caption{Samples from TXUXIv3 dataset~\cite{miro2023assessing} and their respective explanations from a Decision Tree. We can see the sparse nature of these explanations.}
	\label{fig:xai_dt}
\end{figure}

As can be seen in Figure \ref{fig:xai_dt}, where a set of examples of explanations are depicted, the result of this process is a sparse explanation, with a very few pixels with some importance. Therefore, the saliency maps generated from the decision tree model differed significantly from those typically produced by convolutional neural networks (CNNs). This unexpected outcome stems from the fundamental differences between these two types of models. CNNs are based on identifying patterns within specific regions of an image (local patterns), while decision trees favor uncovering relationships across the entire image (global patterns). As a consequence, the decision tree saliency maps do not pinpoint specific, localized regions as important, but instead highlight various pixels throughout the entire image. The algorithm and trained models are publicly available at \url{https://github.com/explainingAI/stability}.

\subsection{Performance measures}

The AI models performance is an important element that can, hypothetically, affect the performance of the robustness metrics. To measure this performance we used standard measures: Mean Absolute Error (MAE), and Mean Squared Error (MSE) (see Equation \ref{eq:mae} and Equation \ref{eq:mse} respectively). These two measurement are the standard de facto for measure the performance of regression problems.

\begin{equation}
    \mae = \frac{\sum^{n}_{i=i} |y_i - \hat{y}_i|}{n},
    \label{eq:mae}
\end{equation}

\begin{equation}
    \mse = \frac{\sum^{n}_{i=i} (y_i - \hat{y}_i)^2}{n},
    \label{eq:mse}
\end{equation}

where $y_i$ is the ground truth and $\hat{y}$ the prediction of the model.

\subsection{Datasets}

We trained the AI model discussed within this section with a simple dataset: TXUXIv3~\cite{miro2023assessing}.

\begin{figure}[htbp]
	\centering
     	\includegraphics[width=0.24\textwidth]{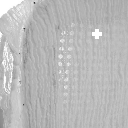}
    	\hfil
    	\includegraphics[width=0.24\textwidth]{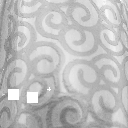}
        \hfil
    	\includegraphics[width=0.24\textwidth]{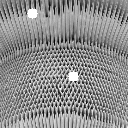}
    	\hfil
        \includegraphics[width=0.24\textwidth]{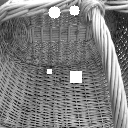} \\ 
     \caption{Samples from TXUXIv3 dataset~\cite{miro2023assessing}, we can see the different background from DTD dataset~\cite{cimpoi14describing}.}
	\label{fig:txuxi}
\end{figure}

TXUXIv3 first introduce by Miró-Nicolau \emph{et al.}~\cite{miro2023assessing} consisted of a set of synthtetic 50000 samples for the training and 2000 for validation. The images are generated combining simple geometric samples (squares, crosses and circles) over a texture background, from the Describable Textures Dataset (DTD)~\cite{cimpoi14describing}. See Figure \ref{fig:txuxi} for examples from this dataset. The original goal of this dataset was to be a Synthetic Attribution Benchmark (SAB), i.e. a dataset containing both ground truth for the prediction task and for the explanation, this is accomplished combining simple images and an attribution function. 

An attribution function is characterized to have the shape from Equation \ref{eq:attr}. This shape is the reason that allowed the SAB datasets know both the prediction GT and the explanation GT.

\begin{equation}
    f(x) = \sum^n_{j=0} w_j \cdot g(p_j, x),
    \label{eq:attr}
\end{equation}

where $p_j$ is the visual pattern $j$, e.g. circles in the image, $x$ is an input image, $g(p_j, x)$ is a function that summarize, numerically, pattern $p_j$ present in image $x$, e.g. the amount of times $p_j$ appeared at image $x$, and finally $w_j$ the weight given to each visual pattern, for the function and therefore the value of the GT explanation itself, as demonstrated by Miró-Nicolau \emph{et al.}~\cite{miro2023novel}. 

We used $ssin$ as the attribution function proposed by Cortez and Embrechts~\cite{CORTEZ20131}, a regression function considering three different elements. See Equation \ref{eq:ssin} for the exact formulation.

\begin{equation}
    \ssin(x) = w_1 \cdot \sin(\frac{\pi}{2} g(p_1, x)) + w_2 \cdot \sin(\frac{\pi}{2} g(p_2, x)) + w_3 \cdot \sin(\frac{\pi}{2} x_3),
    \label{eq:ssin}
\end{equation}

where, $g(p_{i}, x)$ indicates the number of instance of pattern $p_{i}$; and $w_i$ is the weight for each pattern ($w_{1} = 0.55$, $w_{2} = 0.27$, $w_{3} = 0.18$).

This function has one main restriction: the output of the associated function, $g(p_{i}, I)$, must be in the range $[0, 1]$. This range was defined because the maximum value of $\sin$ was obtained using $\pi/2$. Therefore, the maximum value for the $\ssin$ function was obtained when all factors had the maximum value of $1$, essentially when $x_1$, $x_2$ and $x_3$ were equal to $\pi/2$.

\subsection{Experiments}

We realized two different experiments aiming to evaluate the stability metric in different scenarios, using the tests defined in Section \ref{sec:method}.  

\begin{itemize}
    \item \textbf{Experiment 1: Perfect Explanation Test}. We trained a Decision Tree~\cite{breiman1984classification} with the TXUXIv3 dataset~\cite{miro2023assessing} and $\ssin$ attribution function. This experiment allowed us to analyse the behaviour of stability metrics for a regression problem. In addition, using a simple dataset as TXUXIv3 allowed us to have an AI models with good performance.
    \item \textbf{Experiment 2: Random Explanation Test}. For this experiment we evaluate the complementary of the previous experiment: given a random explanation and output for each image any correct stability measure must show the lack of robustness. We generated the explanations sampling from a Gaussian distribution ($\mu = 0, \sigma = 1$) and the prediction value from an uniform distribution, with values between $0$ and $1$. This experiment is inspired by the proposal of Adebayo \emph{et al.}~\cite{adebayo2018sanity}, that randomize element of an XAI pipeline to study the sensibility of XAI methods to this random elements.
\end{itemize}

In each experiment we obtained the performance measures and explanations as explained in this section
\section{Results and discussion} \label{sec:results}

In this section we showed and discussed the results for each experiment defined in the previous section. In both experiments we knew \textit{a priori} the exact robustness value, therefore any metric value a part from this, showed the incorrectness of the measure. {\color{black} This expected value is indicated in all results table to allow a simpler discussion of the results.}

%Table \ref{tab:perform} showed the validation results for each experiment. As indicated in both Section \ref{sec:method} and Section \ref{sec:experimental_setup} due to the transparent nature of the models trained (Decision Trees) we can dispose of a GT for the explanation, hence any result different than perfection, in any experiment, for any measure showed the incorrectness of this measure. 

% \begin{table}[!htb]
% \centering
% \begin{tabular}{lccccc}
% \toprule
% Experiment                          & $\mae$    & $\mse$    & $\precision$  & $\recall$ & $\fscore$ \\
% \midrule
% Exp. 1: TXUXIv3 - $\ssin$       & $0.0333$  & $0.0023$  & -             &  -        &   -           \\
% Exp. 2: TXUXIv3 - $\fnclass$    & -         & -         & $0.72$        & $0.72$    &  $0.72$       \\
% Exp. 3: Caltech101              & -         & -         &               &           &               \\
% Exp. 4:                         & -         & -         & -             &  -        & -             \\
% \bottomrule
% \end{tabular}
% \caption{Validation metrics obtained in the three different experiments. Dashes indicated that because to the experiment nature this metric is not calculated.}
% \label{tab:perform}
% \end{table}

\subsection{Experiment 1: Perfection Explanation Test}

As we already explained in the previous section, we trained a Decision Tree~\cite{breiman1984classification} with the TXUXIv3 dataset~\cite{miro2023assessing} and $\ssin$ attribution function. This task is a regression task. {\color{black} To select the best hyperparameters for this task we realised an exhaustive search over the different values indicated in Table \ref{tab:hyper}. In this table we can also see the best resulting hyperparameter combination in bold. The resulting Decision Tree has $64337$ nodes with a maximum depth of $3064$}. 

\begin{table}[!htb]
\caption{Hyperparameters values used for the training of the Decision Tree. In bold the best combination.}
\centering
\begin{tabular}{ccccc}
\toprule
Criterion        & Splitter      & Max Depth     & Min Sample Split & Max features  \\
\midrule
Squared error    & \textbf{Best} & 7             & 1                & AUTO          \\
Friedmane MSE    & Random        & 30            & \textbf{2}       & SQRT          \\
Absolute error   & -             & 150           & 5                & \textbf{log2} \\
\textbf{Poisson} & -             & 300           & 25               & -             \\
Squared error    & -             & \textbf{None} & 50               & -             \\
-                & -             & -             & 100              & -            \\
\bottomrule
\end{tabular}
\label{tab:hyper}
\end{table}

In Table \ref{tab:perform} we can see validation performance measures. In particular, we can see, that for this experiment, we obtained almost perfect results on both $\mae$ and $\mse$ measures. 

\begin{table}[!htb]
\caption{Validation performance metrics obtained in for Experiment 1}
\centering
\begin{tabular}{lcc}
\toprule
Experiment                      & $\mae$    & $\mse$  \\
\midrule
Exp. 1: Perfection Explanation Test      & $0.033$  & $0.002$  \\       
\bottomrule
\end{tabular}
\label{tab:perform}
\end{table}

In Table \ref{tab:txuxi_ssin} we can see the results of the robustness measures introduced in the previous section, for the trained model: {\color{black} the expected value (for a correct stability measure), the actual value and the Confidence Interval with $0.05$ significance level}. We clearly see that all metrics have obtained perfect results (value equal to $0$), {\color{black} therefore we can assert that both MAX-Sensitivity and AVG-Sensitivity~\cite{yeh2019fidelity} have passed the PET test, without any complex analysis. In this case the results are straightforward, but we can see that the expected value is clearly in between the Confidence Interval, the objective way to evaluate the correctness of these tests.}

\begin{table}[htb]
    \caption{Results from Experiment 1. The table shows the {\color{black} expected perfect value, and the actual value} of the different Robustness measures, {\color{black} in the PET text context,} where the least the better, being $0$ the perfect value.}
    \centering
    \begin{tabular}{lccc}
    \toprule
    Metric                                              & Expected Value    & Actual value          & $CI_{\alpha = 0.05}$       \\ \midrule
    MAX-Sensitivity~\cite{yeh2019fidelity}              & $0.0$             & $0.000 \pm 0.000$     & $(0.0, 0.0)$      \\
    AVG-Sensitivity~\cite{yeh2019fidelity}              & $0.0$             & $0.000 \pm 0.000$     & $(0.0, 0.0)$      \\\bottomrule
    \end{tabular}
    \label{tab:txuxi_ssin}
\end{table}

These results shows that the stability measures analysed worked as expected for a well-trained regression model. In the following experiment we analysed the behaviour of robustness measures in the opposite scenario.

\subsection{Experiment 2: Random Output Test}

This experiment aimed to analyse stability metrics in a completely different context than the previous experiment: instead to use a transparent model that allowed us to have perfect explanation, we will use Gaussian and uniform noise as explanation and predictions respectively, as explained in previous sections. Therefore, any metric indicating a good robustness is erroneous.

\begin{table}[htb]
    \caption{Results from Experiment 2. The table shows the value of the different Robustness measures, where the least the better, being $0$ the perfect value.}
    \centering
    \begin{tabular}{lccc}
    \toprule
    Metric                                     & Expected Value & Actual value          & $CI_{\alpha=0.05}$                                 \\ \midrule
    MAX-Sensitivity~\cite{yeh2019fidelity}     & $1.0$          & $0.011 \pm 0.003$     & $(0.011, 0.011)$                            \\
    AVG-Sensitivity~\cite{yeh2019fidelity}     & $1.0$          & $0.010 \pm 0.003$     & $(0.010, 0.011)$                            \\ \bottomrule
    \end{tabular}

    \label{tab:rnd}
\end{table}

Table \ref{tab:rnd} show the results from this experiment. Due to the usage of noise both as prediction and the explanation any metric result indicating robustness is indicative of an error in the core of the measure. From the results obtained we can see that both Max-Sensitivity~\cite{yeh2019fidelity} and AVG-Sensitivity~\cite{yeh2019fidelity} depict the results as perfect. {\color{black} This large difference between the expected value and the actual value is a clear sign of the incorrectness of these metrics, evermore Confidence Interval, as an objective analysis of the results, showed that the expected value is outside of it.} This error happens due to the nature of both metrics analysed in Section \ref{sec:state-of-the-art}: the fact that in both Equation \ref{eq:sens_max} and Equation \ref{eq:sens_avg} the perturbation samples toke into account are the ones with a prediction difference, respect to the original data, lower than a threshold $\epsilon$, produced an artificial mitigation of the lack of robustness of the methods. 

{\color{black} The two metrics analysed in this experimentation yielded clear results: both have clearly passed the PET test and failed the ROT test. In this case the interpretation is simple because in both the PET and ROT tests the robustness metrics yielded perfect stability, even so we knew that in the second test any correct metric must indicate unrobustness. However, such easy to interpret results may not be typical. We anticipated encountering metrics with less definitive outcomes. Nonetheless, the straightforward nature of the tests, and their simplicity allowed, to expect perfect results for any correct metrics. The expected result will be highly depending on the metric itself, therefore an in-depth analysis of the studied robustness metric must be done to be sure of the correctness of this value, even so the usage of Confidence Intervals would allow a simple way to be sure whether the tests were passed or not.}

\section{Conclusions} \label{sec:concl}

In this study we defined a novel methodology to meta-evaluate the quality of this robustness measures. We defined two new tests: the Perfect Explanation Test (PET) and the Random Output Test (ROT). Both tests are based on, \textit{a priori}, knowledge of the expected robustness results. 

The first test, PET, was based on the usage of a transparent model that allowed us to obtain the real, and perfect, explanation. Therefore we knew that all explanation features should be perfect. Consequently any robustness measure that did not indicate this perfect results showed its flaws. 

The second test, ROT, was the opposite to the PET test: instead of analysing the performance of robustness measures in a perfect context we did it in a completely random behaviour. This random context was achieved with both the explanation and output being randomized. {\color{black} Therefore, any robustness metric with a value different from completely ``unrobust'' do not pass the ROT test.}

We defined two experiments consisting on using these tests to analyse the metrics proposed by Yeh \emph{et al.}~\cite{yeh2019fidelity}: AVG-Sensitivity and MAX-Sensitivity. We used these because their were already analysed in the only previous work that aimed to meta-evaluate stability, Hedström \emph{et al.}~\cite{hedstrom_meta-evaluation_2023}, and due to the similarity to the robustness measure by Alvarez-Melis \emph{et al.}~\cite{Alvarez-Melis2018}, {\color{black} both authors proposed to calculate the robustness using the sensitivity of a function via its gradient $w.r.t.$ of the input}.

The first experiment clearly showed the correct behaviour of both metrics with perfect explanations, with exact values of $0$. However the second experiment showed the inability of both metrics to detect the big lack of robustness from the explanation, with also value very near to $0$ ($0.11$ and $0.010$ respectively). This behaviour is provoked due to the definition of the locality, one of the main features of these measures, that mitigated the lack of robustness of the random explanations. {\color{black} Nonetheless we defined also a methodology to analyse whether a metric pass or not the tests when the results are not so clear as in the case study: we propose to use Confidence Interval and the expected value as a simple and objective way to analyse the test results.}

As future work on the meta-evaluation of robustness measures this work allowed a further comparison with new stability metrics working as an objective benchmark. On the robustness calculation it is clear that the proposal from Yeh \emph{et al.} showed an inherent limitation that made their results unreliable. We hopped that the results obtained trigger novel approaches to the robustness measurement.

\bibliographystyle{splncs04}
\bibliography{references}

\end{document}